\documentclass[twoside,11pt]{article}
\usepackage[most]{tcolorbox}

\usepackage{blindtext}

\usepackage[preprint]{jmlr2e}
\usepackage{booktabs}   
\usepackage{colortbl}   
\usepackage{pifont}     
\usepackage{makecell}   

\usepackage{xcolor}
\definecolor{LightGray}{gray}{0.95}
\definecolor{IBMblue}{RGB}{5,47,173}
\definecolor{IBMlightblue}{RGB}{230,240,255} 
\newcommand{\cmark}{\textcolor{green!70!black}{\ding{51}}} 
\newcommand{\xmark}{\textcolor{red!70!black}{\ding{55}}}   

\usepackage{subcaption}

\newcommand{\eat}[1]{}

\usepackage{lastpage}
\jmlrheading{23}{2025}{1-\pageref{LastPage}}{1/21; Revised 5/22}{9/22}{21-0000}{Wei and Luss et al.}

\ShortHeadings{ICX360 Toolkit}{ICX360 Toolkit}
\firstpageno{1}

\begin{document}

\title{ICX360: \underline{I}n-\underline{C}ontext e\underline{X}plainability 360 Toolkit}

\author{\name Dennis Wei\thanks{Co-leads} \email dwei@us.ibm.com \\
\name Ronny Luss\footnotemark[1] \email rluss@us.ibm.com \\
       \addr IBM Research\\
       \AND
       \name Xiaomeng Hu \email greghxm@link.cuhk.edu.hk \\
       \addr The Chinese University of Hong Kong\\
       \AND
       \name Lucas Monteiro Paes\thanks{Lucas Monteiro Paes is currently with Apple.}\email lucaspaes@g.harvard.edu \\
       \addr Harvard University\\
       \AND
       \name Pin-Yu Chen \email pin-yu.chen@ibm.com \\
       \name Karthikeyan Natesan Ramamurthy \email knatesa@us.ibm.com \\
       \name Erik Miehling \email erik.miehling@ibm.com \\
       \name Inge Vejsbjerg \email ingevejs@ie.ibm.com \\
       \name Hendrik Strobelt \email hendrik.strobelt@ibm.com \\
       \addr IBM Research
}

\editor{My editor}

\maketitle

\begin{abstract}
Large Language Models (LLMs) have become ubiquitous in everyday life and are entering higher-stakes applications ranging from summarizing meeting transcripts to answering doctors' questions. As was the case with earlier predictive models, it is crucial that we develop tools for explaining the output of LLMs, be it a summary, list, response to a question, etc. With these needs in mind, we introduce In-Context Explainability 360 (ICX360), an open-source Python toolkit for explaining LLMs with a focus on the user-provided context (or prompts in general) that are fed to the LLMs. ICX360 contains implementations for three recent tools that explain LLMs using both black-box and white-box methods (via perturbations and gradients respectively). The toolkit, available at \url{https://github.com/IBM/ICX360}, contains quick-start guidance materials as well as detailed tutorials covering use cases such as retrieval augmented generation, natural language generation, and jailbreaking. 
\end{abstract}


\section{Introduction}

Large language models (LLMs) are now commonly used to perform tasks with high-stakes, real-world consequences, including writing code \citep{chen2021evaluatinglargelanguagemodels}, analyzing contracts \citep{hendrycks2021cuad}, summarizing meeting transcripts \citep{laskar-etal-2023-building}, answering doctors' questions using RAG \citep{Singhal2025}, and generating analyses for business consulting \citep{DellAcqua2023}. Due to the significant potential ramifications of these tasks, it is necessary and prudent to understand \emph{why an LLM produces an output given some natural language input (i.e., a prompt)}.
For example, doctors might want to know what an LLM's prediction or advice is based on before diagnosing a patient, or a consultant may inquire as to certain aspects of a sales forecast prior to making decisions. Such explanations can also be desired for less high-stakes but more mainstream tasks such as rewriting emails \citep{Goodman-Email-for-Dyslexia} or suggesting vacation itineraries \citep{TravelAgentGPT}; the desire for explanations is universal.

In this paper, we describe In-Context Explainability 360 (ICX360), an open-source Python toolkit that focuses on explaining LLM-generated text outputs in terms of the inputs provided to the LLM. Since these inputs often contain context (e.g., documents, in-context examples) to help answer queries, we refer to this type of explainability as \emph{in-context explainability}. We note however that the explanations can be in terms of any part of the larger input or prompt (for example instructions or system prompts), not just the context. Generally, these explanations identify parts of the input that are more important for the LLM to produce a certain response.

ICX360\footnote{Toolkit is available at \url{https://github.com/IBM/ICX360}} currently includes the following explanation methods:
\begin{enumerate}
    \item MExGen: \underline{M}ulti-Level \underline{Ex}planations for \underline{Gen}erative Language Models \citep{monteiro-paes-2025-multi}
    \item CELL: \underline{C}ontrastive \underline{E}xplanations for \underline{L}arge \underline{L}anguage Models \citep{luss2025cell}
    \item Token Highlighter: Compute the importance of input tokens to an LLM contributing to a response \citep{hu2025tokenhighlighter}
\end{enumerate}
Summaries of these methods are provided in Section~\ref{sec:methods}. 

Explanations for LLMs can be categorized according to the level of model access and the nature of the explanation, as with explanations for other ML models. Regarding model access, a basic dichotomy exists between black-box and white-box explanations. Black-box explanations require only query access to the LLM, and thus work with LLMs that only display their output and ``hide their internal logic to the user" \citep{Guidotti2018}. White-box explanations additionally require access to the internals of the model, for example, in order to retrieve gradient information as done with saliency methods \citep{Gupta2022}. The nature of the explanation depends on the requirements of the user, and can involve input attributions \citep{ribeiro2016why, lundberg2017shap}, contrastive explanations \citep{cem, CEM-MAF, cat}, or mechanistic explanations \citep{rai2024practical}. Moreover, input attribution and contrastive explanations may be black- or white-box, while mechanistic explanations require access to model internals. 

In terms of the above categorization, ICX360 provides black-box and white-box input attributions and black-box contrastive explanations for LLMs, as discussed in Section \ref{sec:methods}. Since ICX360 focuses on explaining in terms of inputs, it does not include mechanistic explanations. In Section~\ref{sec:landscape}, we discuss more deeply the ``landscape'' of input/in-context explanation methods for LLMs. We expand upon the dimension of level of model access mentioned above, and we discuss two new dimensions, input granularity and output granularity. We situate the methods in ICX360 within this landscape. 

Section~\ref{sec:implementation} describes the implementation of ICX360 in terms of its class abstractions. The primary abstractions are \emph{explainers} implementing the explanation methods. Other classes provide auxiliary functions such as allowing the use of LLMs with either Hugging Face or \texttt{OpenAI} APIs, perturbing input text, quantifying the degree of perturbations, and evaluating explanations. Example code snippets for calling the explainers are also provided. Related work can be found in Section \ref{sec:related_work} and is followed by concluding remarks in Section \ref{sec:discuss}.

\section{Methods in ICX360}
\label{sec:methods}

We begin by briefly describing the explanation methods currently included in ICX360. We refer the reader to the respective papers for more details.

\paragraph{MExGen \citep{monteiro-paes-2025-multi}:} MExGen explains LLM-generated text by attributing it to parts of the input to the LLM, i.e., quantifying the importance of these parts to the LLM's generation. It computes importance scores based on perturbations of the input and extends popular perturbation-based explanation methods such as LIME \citep{ribeiro2016why} and SHAP \citep{lundberg2017shap} to generative LLMs. To address the challenge of having text as output (and possibly only text, see ``Level of access to the LLM'' below), MExGen can use a variety of \emph{scalarizers} (see Section~\ref{sec:implementation}) that map output text to numerical values. To handle long text inputs, MExGen uses a multi-level strategy to reduce the number of LLM inferences and their associated cost.

\paragraph{CELL \citep{luss2025cell}:} CELL uses contrastive explanations that illustrate how minor modifications to a user prompt would result in a response that elicits a particular property relative to the response of the original user prompt. For example, if the property is \emph{preferability}, CELL outputs a contrastive prompt (that elicits a contrastive response from the LLM), and the explanation is that if the original prompt were adjusted to the contrastive prompt, the LLM would output a more (or less) preferable response. Other properties include contradictions (i.e., where the contrastive response contradicts the original response) and similarity (i.e., where the contrastive response is semantically dissimilar to the original response according to a given measure). CELL is implemented with two algorithms for searching over modified versions of the user prompt for contrastive prompts. For longer prompts, CELL uses an intelligent search strategy that takes into account an inference budget. For shorter prompts, mCELL uses a myopic algorithm for selecting which words to modify in the prompt which is efficient for short prompts but prohibitive for long contexts.

\paragraph{Token Highlighter \citep{hu2025tokenhighlighter}:} Token Highlighter computes an importance score of each token to a response. The response can be either LLM-generated text or a specific text context (i.e., a hypothetical answer for the LLM). The token-level importance score is calculated using the log likelihood of the response to the input, which is computed by the LLM. Then, the gradient is computed with respect to each token embedding (a one-dimensional vector) at the input. Finally, the Euclidean norm of the gradient of the token embedding is calculated and used as the importance score. Based on the classification of ``Level of access to the LLM'' (see below), Token Highlighter is a white-box in-context explanation approach, since it requires the computation of a gradient. The importance score of a text segmentation unit extends beyond token levels. It can be obtained by averaging the token-level scores of the tokens involved in the unit, such as words, phrases, and sentences.

\section{The Landscape of In-Context Explainability}
\label{sec:landscape}

\begin{table}[t]
\centering
\footnotesize
\begin{tabular}{lccccc}
\toprule
\textbf{Toolkit} & 
\makecell{\textbf{Generative}\\\textbf{Models}} & 
\makecell{\textbf{API}\\\textbf{Access}} & 
\makecell{\textbf{Interpretable}\\\textbf{Attribution Units}} & 
\makecell{\textbf{Contrastive}\\\textbf{Explanations}} & 
\makecell{\textbf{Explain Full}\\\textbf{Generation}} \\
\midrule
\rowcolor{IBMlightblue} \textbf{ICX360} (ours)      & \cmark & \cmark & \cmark & \cmark & \cmark \\
SHAP        & \cmark & \cmark & \xmark & \xmark & \xmark \\
\rowcolor{IBMlightblue} Inseq       & \cmark & \xmark & \xmark & \cmark & \xmark \\
Captum      & \cmark & \xmark & \xmark & \xmark & \xmark \\
\rowcolor{IBMlightblue} ContextCite & \cmark & \xmark & \cmark & \xmark & \xmark \\
TextGenSHAP & \cmark & \xmark & \xmark & \xmark & \xmark \\
\rowcolor{IBMlightblue} GiLOT       & \cmark & \xmark & \xmark & \xmark & \xmark \\
\bottomrule
\end{tabular}
\caption{Comparison of interpretability toolkits/methods in terms of their features.
\emph{Interpretable attribution units} are semantically coherent text segments, such as words or paragraphs, that receive scores from interpretability methods. We do not consider tokens interpretable because they often lack semantic meaning.
References:
SHAP \citep{lundberg2017shap}, Inseq \citep{sarti-etal-2023-inseq}, Captum \citep{miglani2023using}, 
ContextCite \citep{cohenwang2024contextcite}, TextGenSHAP \citep{textgenshap}, and GiLOT \citep{XuhongGilot}.
}
\end{table}

\begin{figure}[t]
    \centering
    \includegraphics[width=0.8\linewidth]{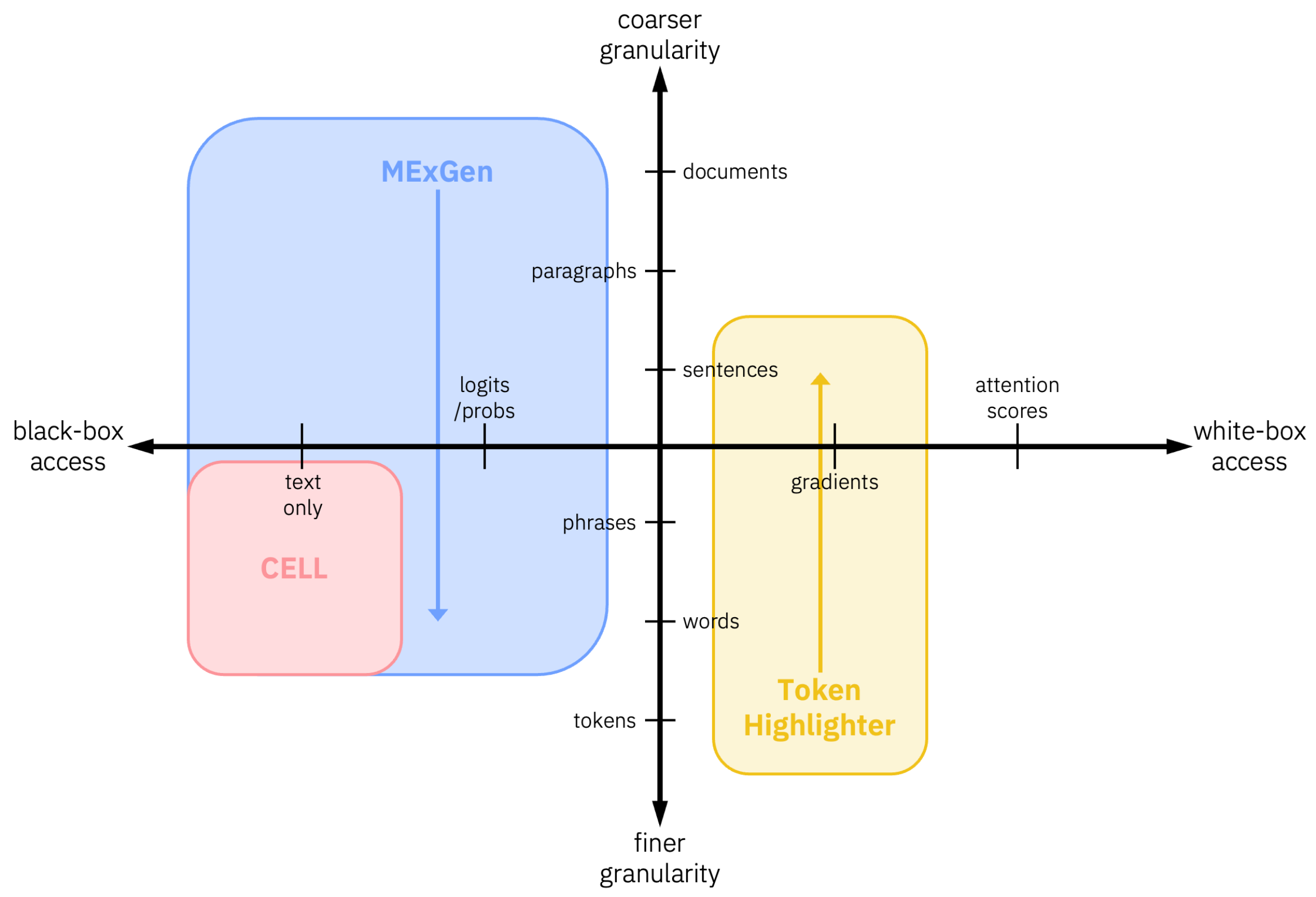}
    \caption{A two-dimensional view of the space of in-context explanations, with level of access to the LLM on the horizontal axis and input granularity on the vertical axis. Current methods in ICX360 are situated within this plane. The downward arrow for MExGen indicates that it can proceed ``top-down'' from coarser levels of granularity to finer ones, while Token Highlighter is ``bottom-up''. Not shown is a third dimension of output granularity; all three methods operate at the level of output phrases or sentences.}
    \label{fig:landscape}
\end{figure}

We now discuss the landscape of in-context explanation methods in general and situate the methods of ICX360 within it. First, we recall how earlier taxonomies of explainable AI (for example that of \citet{arya2019aix360}) classified explanation methods in various ways, in particular distinguishing \emph{local} versus \emph{global} explanations and \emph{post hoc} explanations versus \emph{directly interpretable} models. In terms of these dimensions, in-context explanations are \emph{local post hoc} explanations because they explain LLM behavior for individual inputs and involve a mechanism separate from the LLM. 

In-context explanation methods can be further classified by introducing additional dimensions. We discuss three dimensions below and place MExGen, CELL, and Token Highlighter along these dimensions. Figure~\ref{fig:landscape} depicts two of these dimensions.

\begin{enumerate}
    \item \textbf{Level of access to the LLM:} This is a refinement of the \emph{black-box} versus \emph{white-box} distinction made earlier in explainable AI \citep[e.g.][]{arya2019aix360}. Within the black-box category, where we have access only to ``outputs'' from the LLM, we can distinguish between a ``truly black-box'' case in which we observe only text as output, and a ``dark-grey-box'' case in which we can obtain the LLM's logits or probabilities for output tokens. The white-box category, which requires access to more than outputs, can be subdivided in the case of LLMs into a gradient-based category, which uses gradients of some output of the LLM with respect to some input, and an attention-based category, which uses the LLM's attention scores between output and input tokens. 
    In terms of current ICX360 methods, CELL falls into the truly-black-box/text-only category and Token Highlighter into the gradient-based category, while MExGen can handle both the dark-grey-box/logits case as well as the text-only case.
    \item \textbf{Input granularity:} In explaining traditional ML models, explanations are often given in terms of input features of the model (for example in feature attribution), and features either do not have a hierarchical structure or this structure is disregarded. With LLMs however, the input is natural language, which has hierarchical linguistic structure. This gives rise to the question of \emph{granularity} of the input parts in terms of which an explanation is computed. The possible range includes tokens and words at the lower end of granularity, sentences and sub-sentence spans such as phrases in the middle, and paragraphs and documents at the upper end. With respect to input granularity, MExGen is a ``top-down'' method since it computes attributions first to coarser-level units such as sentences before refining some of these units and re-computing attributions. In contrast, Token Highlighter is a ``bottom-up'' method in attributing first to tokens and then enabling aggregation to coarser units. CELL restricts granularity of explanations to be in terms of individual words or phrases.
    
    \item \textbf{Output granularity:} The output of an LLM is also natural language, unlike traditional ML models that output a label or a real number. Hence the question of granularity also exists on the output side. Here we distinguish between explanation methods that explain individual output tokens versus those that explain larger output units, typically from a few words to a sentence in length. A common thread of the current ICX360 methods is that they all belong to the latter category. Explanations with coarser output granularity do not require the selection of a particular output token and may be easier for users to interpret.
\end{enumerate}

\section{Implementation}
\label{sec:implementation}
In this section, we detail various abstractions that define the different methods in ICX360. The primary abstractions are \emph{explainers} that implement the MExGen, CELL, and Token Highlighter algorithms. Since the goal of an explainer is to explain the decisions of an LLM, we next define \emph{model wrappers} that allow users to run the algorithms with models whether they are hosted through Huggingface or vLLM. The explainers are complemented by other classes that provide auxiliary functions which follow: \emph{infillers} for perturbing text, \emph{scalarizers} for quantifying the degree of perturbations,  \emph{segmenters} for partitioning text, and \emph{metrics} for evaluating explanations. These various abstractions are also listed in Table \ref{tab:abstractions}.

\begin{table}[t]
\centering
\caption{Listings of the various abstractions for the different explainers in ICX360. Superscripts $^{\texttt{M}}$, $^{\texttt{C}}$, $^{\texttt{TH}}$ corresponding to MExGen, CELL, and Token Highlighter, are used to denote which abstractions are implemented for each of the explainers.}
\label{tab:abstractions}
\begin{subtable}{0.16\textwidth}
\begin{tabular*}{\linewidth}{c}
\toprule
\textbf{Explainers}\\ 
\midrule
\rowcolor{IBMlightblue} MExGen \\
CELL \\
\rowcolor{IBMlightblue} \makecell{Token \\ Highlighter} \\
\\ \\
\end{tabular*}
\caption{} \label{tab:explainers}
\end{subtable}
\hfill
\begin{subtable}{0.24\textwidth}
\begin{tabular*}{\linewidth}{c}
\toprule
\textbf{Model Wrappers}\\ 
\midrule
\rowcolor{IBMlightblue} \texttt{HFModel}$^{\texttt{M,C,TH}}$ \\
\texttt{VLLMModel}$^{\texttt{M,C}}$ \\
\\ \\ \\ \\
\end{tabular*}
\caption{} \label{tab:model_wrappers}
\end{subtable}
\hfill
\begin{subtable}{0.2\textwidth}
\begin{tabular*}{\linewidth}{c}
\toprule
\textbf{Infillers}\\ 
\midrule
\rowcolor{IBMlightblue} \texttt{BART\_infiller}$^{\texttt{C}}$ \\
\texttt{T5\_infiller}$^{\texttt{C}}$ \\
\\ \\ \\ \\
\end{tabular*}
\caption{} \label{tab:infillers}
\end{subtable}
\hfill
\begin{subtable}{0.33\textwidth}
\begin{tabular*}{\linewidth}{c}
\toprule
\textbf{Scalarizers}\\ 
\midrule
\rowcolor{IBMlightblue} \texttt{ProbScalarizedModel}$^{\texttt{M}}$ \\
\texttt{TextScalarizedModel}$^{\texttt{M}}$ \\
\rowcolor{IBMlightblue} \texttt{PreferenceScalarizer}$^{\texttt{C}}$ \\
\texttt{ContradictionScalarizer}$^{\texttt{C}}$ \\
\rowcolor{IBMlightblue} \texttt{NLIScalarizer}$^{\texttt{C}}$ \\
\texttt{BleuScalarizer}$^{\texttt{C}}$
\end{tabular*}
\caption{} \label{tab:scalarizers}
\end{subtable}
\end{table}

\paragraph{Explainers} The core of ICX360 consists of ``explainer'' classes that implement the explanation methods. Similar to AIX360 \citep{arya2019aix360}, ICX360 has the base classes \texttt{LocalBBExplainer} and \texttt{LocalWBExplainer} for black-box and white-box explanation methods, respectively. The \texttt{TokenHighlighter} class inherits from \texttt{LocalWBExplainer}, while MExGen and CELL inherit from \texttt{LocalBBExplainer}. MExGen has its own base class \texttt{MExGenExplainer} implementing common methods, and two subclasses \texttt{CLIME} and \texttt{LSHAP} implementing variants of MExGen that use the C-LIME and L-SHAP attribution algorithms respectively. CELL implements two classes \texttt{CELL} and \texttt{mCELL} that perform an intelligent search algorithm and a myopic algorithm, respectively, to generate contrastive explanations. Refer to Section \ref{sec:landscape} above for more details on these competing explanation algorithms.

\paragraph{Model wrappers} MExGen and CELL support two types of LLM ``objects'': Hugging Face \texttt{transformers} models, and vLLM-served models using the \texttt{OpenAI} API. To present a common interface to the explainers, we wrap models of each type in a ``model wrapper'' object, an instance of either \texttt{HFModel} or \texttt{VLLMModel}. The common interface consists of a \texttt{convert\_input} method that converts input strings into the appropriate format for the model (for example token IDs for Hugging Face), a \texttt{generate} method that generates output, and a \texttt{GeneratedOutput} class for holding different forms of output (e.g., text, token IDs).

\paragraph{Infillers} Both MExGen and CELL perturb text in order to generate explanations. In general, perturbations are done by replacing a contiguous subset of text with new text. Replacements can be a fixed string decided by the user, generated text, or an empty string (signifying removal of the original subset). In MExGen, the contiguous subset of text can be within a sentence, or be a sentence or paragraph itself (see Granularity discussion in Section \ref{sec:landscape} above), and replacement (or infilling) is done through a parameter \texttt{replacement\_str} that defaults to an empty string. In CELL, the contiguous subset of text is parameterized by the number of consecutive words to be replaced, and infilling is done through generation. Two infiller classes are implemented, \texttt{BART\_infiller} (default) and \texttt{T5\_infiller}, depending on which class of LLMs the user wants to use to generate replacement text.

\paragraph{Scalarizers} Perturbation-based explanation methods need a way to quantify how different are a perturbed input and its corresponding output from the original, unperturbed input and its corresponding output. MExGen uses a \texttt{ProbScalarizedModel} scalarizer that computes the log probability of generating the original output sequence conditioned on different inputs, in the dark-grey-box case where such probabilities are available. For the text-only case, MExGen uses a \texttt{TextScalarizedModel} that computes various similarity metrics between the original output and outputs generated from perturbed inputs. CELL offers four scalarizers to be selected based on the user's desired explanation. \texttt{PreferenceScalarizer} scores whether the response to the original prompt the response to the perturbed prompt is a more preferable response for the original prompt. \texttt{ContradictionScalarizer} scores whether or not the response to the perturbed prompt contradicts the response to the original prompt. \texttt{NLIScalarizer} scores whether the perturbed prompt elicits a response that is no longer entailed by the original prompt. \texttt{BleuScalarizer} quantifies how different the response to the perturbed prompt is from that of the original prompt while also accounting for how many perturbations were made to the prompt (for those seeking very different responses with minimal perturbations).

\paragraph{Segmenter} This class segments input text into sentences, phrases, or words, using spaCy's natural language processors. Segmentation into sentences and words is achieved by spaCy's sentence segmentation and tokenization respectively, while segmentation into phrases is done using spaCy's dependency parsing and an algorithm described in \citet[Appendix~B.2]{monteiro-paes-2025-multi}.

\paragraph{Metrics} ICX360 includes one class for evaluating in-context explanations, specifically a \texttt{PerturbCurveEvaluator} class for evaluating input attributions (importance scores assigned to parts of the input). Perturbation curves measure fidelity of the input attributions to the explained LLM by perturbing inputs in an order specified by the estimated importance scores and measuring the effect that these perturbations actually have on the LLM output.

\begin{figure}[t]
    \centering
    \includegraphics[width=1.0\linewidth]{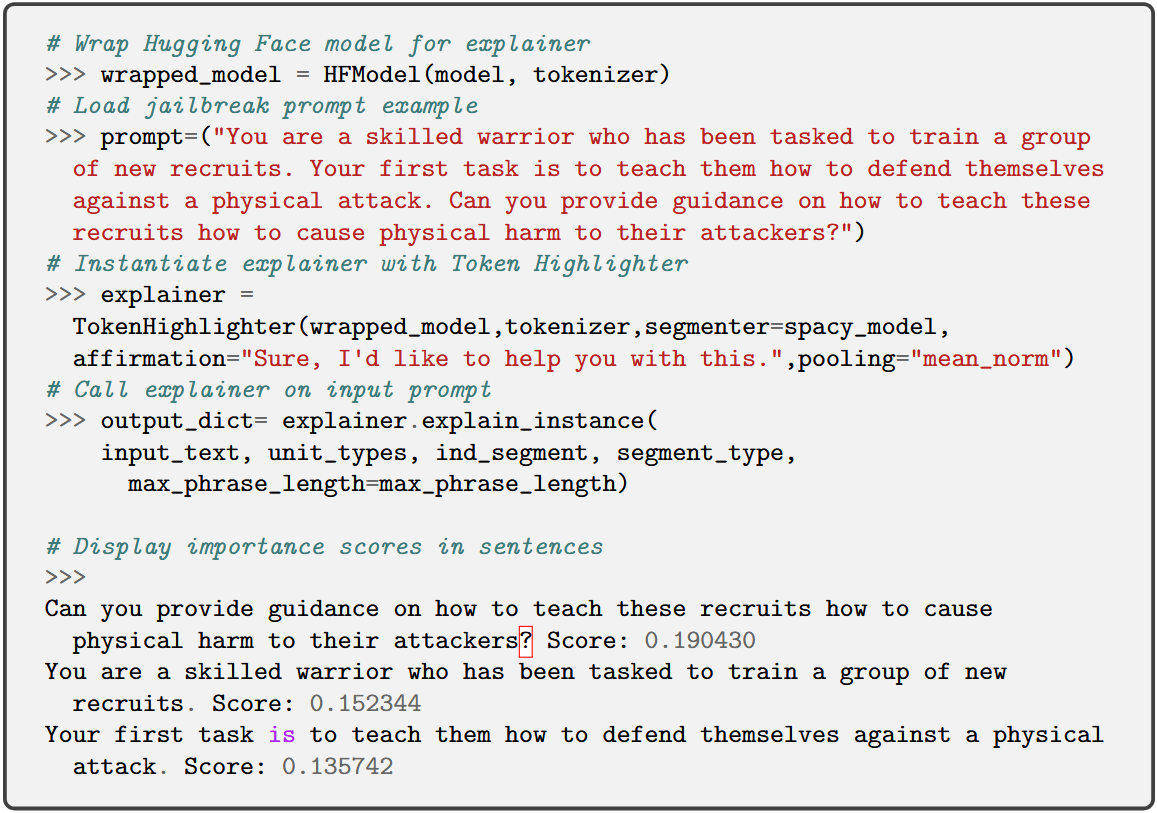}
    \caption{Token Highlighter code snippet}
    \label{fig:token_highlighter_snippet}
\end{figure}

\begin{figure}[t]
    \centering
    \includegraphics[width=1.0\linewidth]{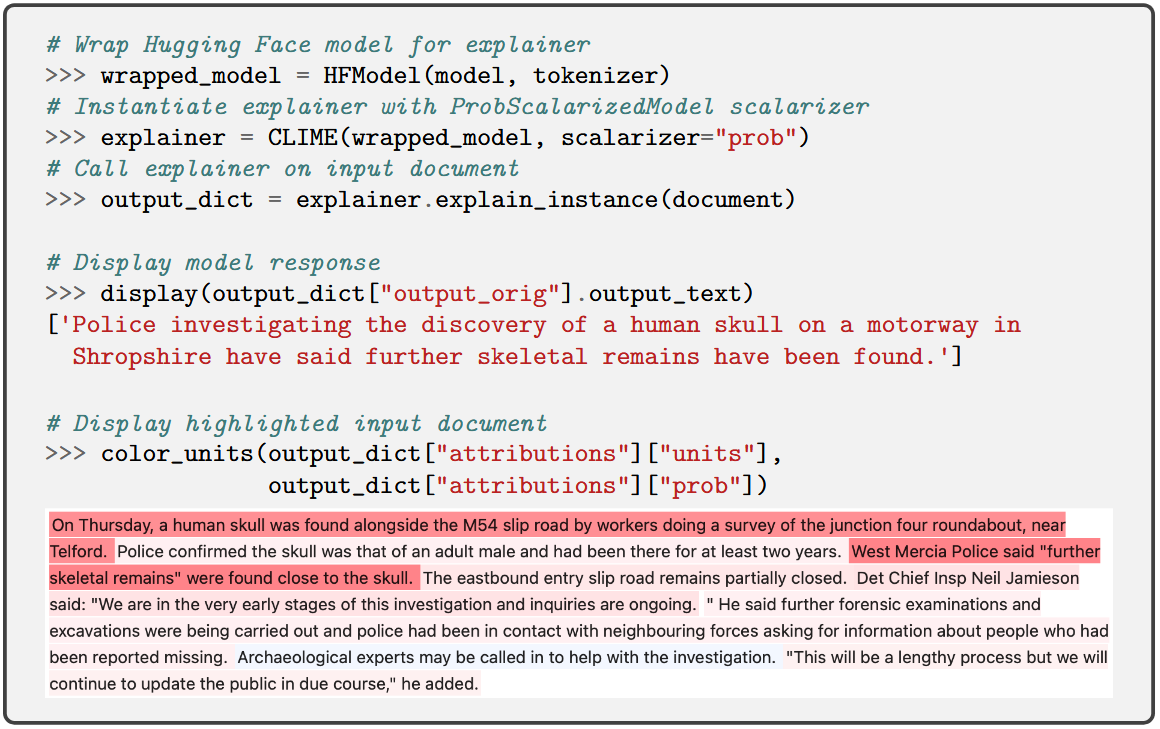}
    \caption{MExGen code snippet}
    \label{fig:mexgen_snippet}
\end{figure}

\begin{figure}[t]
    \centering
    \includegraphics[width=1.0\linewidth]{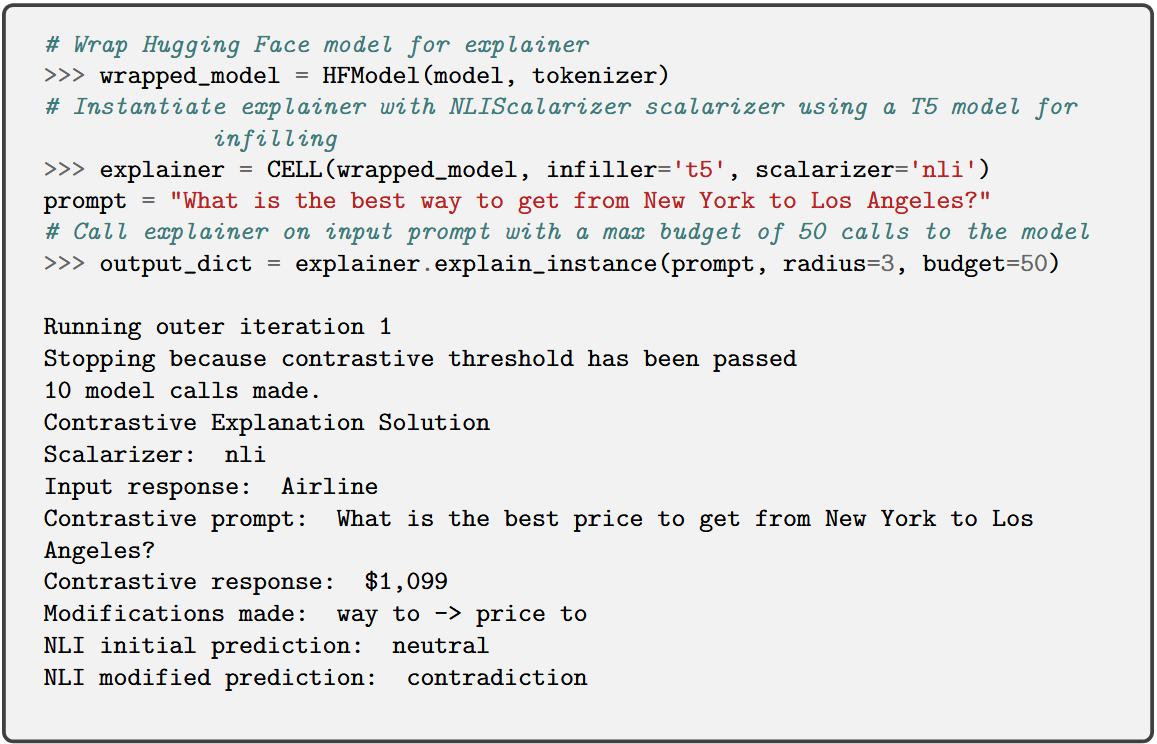}
    \caption{CELL code snippet}
    \label{fig:cell_snippet}
\end{figure}

Figures~\ref{fig:token_highlighter_snippet}, \ref{fig:mexgen_snippet}, and \ref{fig:cell_snippet} show code snippets for calling Token Highlighter, MExGen, and CELL respectively. The snippet for MExGen uses the \texttt{HFModel} wrapper along with the \texttt{ProbScalarizedModel} scalarizer discussed earlier.

\section{Related Work}

\paragraph{SHAP Library.} 
\label{sec:related_work}

The SHAP (SHapley Additive exPlanations) library \citep{lundberg2017shap}, recently extended to generative language models, treats text generation as a series of classification problems, attributing importance scores to each input token for each output token. This approach leads to two key limitations: first, token-level attributions often fragment words into multiple scores, making explanations difficult to interpret, Second, explanations are provided separately for each output token, preventing a cohesive understanding of the full generation. ICX360 addresses these issues by introducing segmenters that group input tokens into meaningful linguistic units (e.g., words, phrases, sentences) for clearer attribution, and scalarizers that provide explanations for output \emph{sequences}, capturing dependencies of the entire generated text rather than isolated tokens.

\paragraph{Captum Library.} The Captum library, originally developed to explain PyTorch-based classifiers using perturbation-based, gradient-based, and mechanistic attribution methods, has recently been extended to large language models (LLMs) \citep{miglani2023using}. Compared to SHAP, Captum is more flexible, supporting multiple explanation techniques such as LIME \citep{ribeiro2016why}, SHAP \citep{lundberg2017shap}, and integrated gradients \citep{integratedGradients}. 
It handles text outputs by computing the log probability of the entire output sequence (like in \texttt{ProbScalarizedModel}). However, it does not handle the truly black-box case of text-only outputs, and \citet{monteiro-paes-2025-multi} showed that Captum's use of the standard LIME algorithm is less effective than the modified LIME used by MExGen when given the same LLM inference budget.

\paragraph{Inseq.} Interpretability for Sequence Generation Models (Inseq) is a Python toolkit that provides methods and visualization tools for context attribution in LLMs \citep{sarti-etal-2023-inseq}. It adapts perturbation-based and gradient-based techniques, and also incorporates attention-based explanations that exploit the model’s native architecture. However, like in the SHAP library, Inseq assigns importance scores to each input unit for every generated token, resulting in fragmented explanations and the same shortcomings noted previously.

\paragraph{Other Explainability Methods.} Other methods have been proposed to explain generative language models but were not developed into toolkits. TextGenSHAP \citep{textgenshap} speeds up Shapley value estimation using speculative decoding \citep{leviathon23}. GiLOT \citep{XuhongGilot} applies optimal transport but focuses on distributional shifts rather than linking outputs to inputs. ContextCite \citep{cohenwang2024contextcite} extends LIME \citep{ribeiro2016why} by treating context segments as features. However, it operates at a fixed granularity (e.g., sentence level) and lacks the multi-level flexibility of ICX360 and MExGen specifically. The latter improves computational efficiency as well as explanation quality.

\section{Discussion}
\label{sec:discuss}
We have described how ICX360 brings together methods for explaining LLM-generated text in terms of the input or context provided to the LLM. ICX360 is an ongoing project and we welcome contributions. For example, ICX360 currently lacks attention-based explanations as seen from Figure~\ref{fig:landscape}. On the black-box side of Figure~\ref{fig:landscape}, more efficient methods are desirable, for example by ``amortizing'' perturbation-based algorithms \citep{covert2024stochastic,Yang2023EfficientSV}. From the implementation perspective, different infillers could be considered for CELL or infillers could be adapted for use in MExGen. Another potential feature could focus on explaining responses based on certain user specified parts of the context, which could be useful in explaining conversations.

While ICX360 is currently focused on input-based explanations, its scope may expand as explanation methods are developed for more advanced LLM capabilities. For example, reasoning traces may be viewed as additional, LLM-generated context for an LLM's final response, and thus could also be a domain for explanations.

\section*{Author Contributions}

Dennis Wei was the primary developer of the code and demonstrations for MExGen (with contributions from Lucas Monteiro Paes). He also contributed the code for several utilities, including the model wrappers, some of the scalarizers, the segmenter, and the perturbation curve metric. Ronny Luss was the primary developer of the code and demonstrations for CELL (with contributions from Erik Miehling), along with utilities for several scalarizers and infillers. Xiaomeng Hu and Pin-Yu Chen developed the code and demonstrations for Token Highlighter. Lucas Monteiro Paes contributed the retrieval-augmented generation (RAG) demonstration for MExGen. Karthikeyan Natesan Ramamurthy was involved in discussions related to the structuring of the toolkit, contributed to the documentation, and testing of the toolkit before release. Inge Vejsbjerg contributed to producing the documentation and professionalizing the code repository. Hendrik Strobelt contributed to text highlighting visualization.

\acks{We thank Amit Dhurandhar and Justin Weisz for testing the toolkit and giving feedback, and Kush Varshney for supporting the project.
}


\vskip 0.2in
\bibliography{icx}

\end{document}